\def\year{2021}\relax
\documentclass[letterpaper]{article} % DO NOT CHANGE THIS
\usepackage{aaai21}  % DO NOT CHANGE THIS
\usepackage{times}  % DO NOT CHANGE THIS
\usepackage{helvet} % DO NOT CHANGE THIS
\usepackage{courier}  % DO NOT CHANGE THIS
\usepackage[hyphens]{url}  % DO NOT CHANGE THIS
\usepackage{graphicx} % DO NOT CHANGE THIS
\usepackage{natbib}  % DO NOT CHANGE THIS AND DO NOT ADD ANY OPTIONS TO IT
\usepackage{caption} % DO NOT CHANGE THIS AND DO NOT ADD ANY OPTIONS TO IT
\usepackage{times}
\usepackage{graphicx}
\usepackage{latexsym}

\usepackage{url}            % simple URL typesetting
\usepackage{booktabs}       % professional-quality tables
\usepackage{amsfonts}       % blackboard math symbols
\usepackage{nicefrac}       % compact symbols for 1/2, etc.
\usepackage{microtype}      % microtypography
\usepackage{amsmath}		% Can be removed after putting your text content
\usepackage{verbatim}
\usepackage{booktabs}
\usepackage{multirow}
\usepackage{hhline}
\usepackage{ulem}
\usepackage{breqn}
\usepackage{adjustbox}
\usepackage{tabularx}
\usepackage[colorinlistoftodos]{todonotes}

\title{Guiding Graph Embeddings using Path-Ranking Methods for Error Detection in noisy Knowledge Graphs}
\author{
    Konstantinos Bougiatiotis\textsuperscript{\rm 1}\textsuperscript{\rm 2}, Romanos Fasoulis\textsuperscript{\rm 1}\footnote{Equal contribution with first author}, Fotis Aisopos\textsuperscript{\rm 1},  Anastasios Nentidis\textsuperscript{\rm 1}, Georgios Paliouras\textsuperscript{\rm 1}\\
}
\affiliations{
    \textsuperscript{\rm 1}Institute of Informatics and Telecommunications, National Center Scientific Research Demokritos, Athens, Greece\\
    \textsuperscript{\rm 2}Department of Informatics and Telecommunications, National and Kapodistrian University of Athens, Greece\\
    kbogas@di.uoa.gr

}
\begin{document}

\maketitle

\begin{abstract}
Nowadays Knowledge Graphs constitute a mainstream approach for the representation of relational information on big heterogeneous data, however, they may contain a big amount of imputed noise when constructed automatically. To address this problem, different error detection methodologies have been proposed, mainly focusing on path ranking and representation learning. This work presents various mainstream approaches and proposes a hybrid and modular methodology for the task. We compare different methods on two benchmarks and one real-world biomedical publications dataset, showcasing the potential of our approach and providing insights on graph embeddings when dealing with noisy Knowledge Graphs.  
\end{abstract}

\section{Introduction}

	A Knowledge Graph (KG) is a construct for representing relational information between entities that can be extracted either manually or automatically, e.g. from text found online. Each piece of information is usually presented as a triple $(s, r, o)$, where $s$ is the subject, $o$ is the object, and $r$ is the relation connecting them. Every such triple is also called a fact~\cite{survey}.
	%, as every triple indicates that two entities are joined by a specific relation %

	In the last decade, as %automatic tools for extracting information and %
	the Natural Language Processing (NLP) domain is growing rapidly, we have seen a surge of growth in automatic knowledge graph construction and development. Knowledge graphs like DBpedia~\cite{DBPedia}, YAGO~\cite{YAGO}, NELL~\cite{NELL} and Wikidata ~\cite{Wikidata}  are automatically created. While the aforementioned knowledge graphs are bigger and more detailed than ever before, it is apparent that errors and noise cannot be avoided, as automatic extraction tools are not perfect. Consequently, it is important to estimate how much noise exists in automatically constructed KG's, how prominent it is and how this noise affects any further downstream tasks that will be performed on the KG (e.g. link prediction or node classification).
	
	To address the problem of noise and errors in knowledge graphs, we focus on contrasting various methods that use different techniques concerning error detection. Although many papers focus on KG completion and link prediction tasks, few actually deal with the problem of existing noise. % in automatically constructed KG's.
	
	In this work, our goal is to compare and combine different approaches, mainly stemming from path ranking analysis and graph embeddings. In addition, we propose a hybrid of the two approaches. %We verify the effectiveness of each method on two benchmark knowledge graphs and a knowledge graph created from a real-world application on biomedical publications.%
	We showcase that a specific combination of a path ranking algorithm and graph embeddings can achieve better results on error detection tasks, while also creating robust-to-noise embeddings that can later be used for further analysis and tasks. 

    The contributions of this work are summarized as follows:
    \begin{itemize}
     \item Development of a hybrid error detection approach that extends the use of path ranking algorithms to generate error-robust embeddings.
     \item Quantitative and qualitative comparison and assessment of different error detection methods based on path ranking, graph embedding and hybrid techniques.
     \item A framework that allows the combination of different error detection tools and graph embedding techniques to tackle challenges in noisy knowledge graphs. 
    \end{itemize}
    
    The rest of the paper is structured as follows: first, related work on the subject is presented and then the details of the proposed hybrid approach are  described. Following that, the experimentation process is presented, alongside comments on the qualitative and quantitative results. Finally, the limitations and conclusions of the current work are discussed.

\section{Related Work}\label{sec:rel_work}

	Error detection tasks in knowledge graphs become more and more prominent as modern, automated ways of constructing knowledge graphs create higher demands regarding data integrity~\cite{Wiki_vandalism}. The quality assessment of data residing in such graphs may be applied in a static way, or taking into account their evolution over time~\cite{rashid2019quality}. There are a handful of methods for error detection in knowledge graphs via fact checking~\cite{shi2016discriminative} or anomaly detection \cite{akoglu2015graph}, with each one targeting various types of information~\cite{KGR}. This information can be internal and present in knowledge graph (e.g. density, structure, etc.) or external (e.g. textual information). 
	
\subsection{Internal and External Error Detection Methods}

	A good example of an internal method is SDValidate~\cite{SDValidate}, which uses the characteristic distribution of types (concepts) and relations. The work in \cite{jia2019triple} presents a knowledge graph triple confidence assessment model, quantifying their semantic correctness based on a Reachable paths inference algorithm. 
	
	In our work,  we employ PaTyBRED~\cite{PaTyBRED}, a path ranking approach that utilizes the Path Ranking Algorithm ~\cite{PRA} and the Sub-graph Feature Extraction ~\cite{SFE} methods for error detection. Recent works showed that the presence of relations between candidate pairs, exploited in path ranking, can be an extremely strong signal of a correct link \cite{toutanova2015observed} and also that a simple method can achieve state-of-the-art performance when well tuned \cite{kadlec2017knowledge}. This is also validated in our results, illustrating the high performance of a simple path ranking approach as will be discussed. PaTyBRED discovers complex patterns over paths between nodes that are used as features in a Random Forest classifier, in order to predict confidence scores for all triples. The feature extraction process of this method is further explained in Methodology. %Additionally, they introduce a K-best selection method before training the classifiers and some heuristic measures for quicker and more robust feature extraction. 

	On the contrary, methods like DeFacto~\cite{Defacto}, which specialize in finding erroneous relations, uses external information in the form of lexicalizations. In this work, we are focusing solely on internal methods.

	%The Path Ranking Algorithm (PRA)~\cite{PRA} is a method that can discover complex patterns in relational data, applying logistic regression over paths between nodes that are used as features. These paths are extracted through feature selection (random walks over the graph). For each triple in the KG, each path is assigned a weight that reflects the probability of arriving to the triple's targeted object, given the triple's subject and the path.
	
	%Sub-graph Feature Extraction (SFE) is an improvement of PRA, proposed by Gardner and Mitchell~\cite{SFE}, aiming to reduce overall complexity, run-time and achieve statistical superiority. Novel improvements in SFE comprise the replacement of PRA's path probability with binary values that reflect the ability to go to the triple's object from the triple's subject, as well as the replacement of the Random Walks method  with a Breadth-First Search (BFS) algorithm. 

\subsection{Embedding Methods}

	While triples are considered to be a very effective structural representation of KGs, the need to better manage and gain access to underlying symbolic information of these triples led to the representation known as knowledge graph embeddings~\cite{survey2}. This involves the transformation of entities and relations of knowledge graphs into lower-dimensional, fixed-size vectors. These vectors can afterward be employed for further downstream tasks such as link prediction, node classification, as well as error detection.

    Baseline approaches on error detection employ graph embeddings based on Translational Distance Models like TransE~\cite{transe} or Semantic Matching models~\cite{survey}, aiming at identifying erroneous triples through the increased values of specified loss functions. The authors of TransT~\cite{zhao2019embedding} introduce a novel translating embedding approach with triple trustfulness, taking low trust (possible noise) triples into the consideration of the energy function, to extract a confidence for all graph triples. On the other hand, the authors in~\cite{abedini2020correction} divide the errors in a Knowledge Graph into three types (outliers, inconsistencies and erroneous relations) and employ Semantic Matching models embeddings in three different trays for the Graph correction.

	\subsection{Rule-based Anomaly Detection}

    To improve the embedding creation process, the work of~ \cite{guo2017knowledge}  proposed an iterative guidance from soft rules, resulting in a Rule-Guided Embedding model (RUGE). The RUGE model has been employed by \cite{hong2020rule} to produce rule-enhanced noisy embeddings for low-quality error detection, focusing on entity type errors. Rule-based error detection has been also performed in~\cite{cheng2018rule} introducing a set of automatic repairing semantic patterns for simple graphs, called Graph-Repairing Rules (GRRs), which focus on conflicts and redundancies, in terms of error detection. Ho et al.~\cite{ho2018rule} worked in the opposite direction, using embeddings to guide rule learning, while Zhang et al. ~\cite{zhang2019iteratively} operated iteratively, so the embeddings and the rules can guide each other.

    Most importantly, \cite{belth2020normal} presented an interesting unified graph summarization and refinement approach, developing a set of soft rules in the form of graph patterns, in order to identify various anomaly types. The resulting model dictated inductive rules for relations between node labels and outperformed baseline embedding methods used for error detection, such as ComplEx and TransE. Our approach, as will be explained, combines embeddings with path ranking, enabling the exploitation of features like frequency of big-length paths, that graph rules may not capture.

\subsection{Embeddings Guided by the Confidence of Triples}

	%It can be argued that all of the embedding methods mentioned in the previous section use topological characteristics of the KG to construct the embeddings of its entities and relations. However, there exist approaches that utilize additional information regarding the entities and the relations found in the KG, to improve performance and expressiveness. Here we will present some of these approaches that have been used in error detection tasks.
	
	Embedding approaches may be enriched with additional internal information of the KG such as the collection of paths connecting a subject to an object. For example, the PTransE model extends TransE, using paths in addition to relations~\cite{ptranse}. %It replaces the relation in the Transitional Requirement with each path connecting a subject and the object, creating many more Translational Requirements to be satisfied and energy functions to be minimized. 
	Similarly, the Confidence-aware KRL framework (CKRL)~\cite{CKRL} introduces a triple confidence score that guides the loss function to pay attention to more "convincing" triples. This confidence score takes into account different aspects and characteristics of triples, both local and global. The Triple trustworthiness measurement model for Knowledge Graph (KGTtm)~\cite{kgttm} uses a crisscrossed neural network-based structure, combining different elements through a multi-layer perceptron fusioner to generate confidence scores for each triple. %The method is tested on the task of error detection, however, errors are not imputed in KG as noise, but are pre-labeled as negatives.%
	
	%Other methods may use other kinds of external information to guide the embeddings. TRESCAL extends the RESCAL model by employing additional external information like entity types/concepts, range and domain restrictions to improve performance~\cite{rescal}. The KALE~\cite{kale} and RUGE~\cite{RUGE} models, as well as the models proposed in~\cite{rules2} and~\cite{rules1} use logic rules and Horn clauses to guide embeddings creation and optimization. In contrast to external resources like textual information, rules are internal information, therefore, always available with extraction tools like AMIE+~\cite{AMIE}. 

    The current approach follows a similar principle, by employing a simpler path ranking approach to guide the construction of knowledge graph embeddings.

\section{Methodology}\label{sec:meth_empl}

	In this section, we  briefly present two methods that will be used later as modular components in the proposed Path Ranking Guided Embeddings (PRGE) approach. Then, we present PRGE as a framework along with the specific model tested here. In the following, a knowledge graph $G$ will be defined as a set of triples. Each triple follows the form of $(s, r, o)$, where $(s, o) \in E$ are the entities and $r \in R$ is the relation that binds them. $E$ is the set that contains all entities that exist in $G$ and $R$ is the set that contains all relations. We assume that $G$ also contains some ratio of noise $N$\%, denoting that $N$\% of the triples in $G$ are erroneous. The objective of error detection  is to find a way to pinpoint these errors in $G$.

\subsection{PaTyBRED}
	
	PaTyBRED~\cite{PaTyBRED} is a path-ranking based algorithm that was developed with the task of error detection in mind. PaTyBRED uses paths as features, with a path being defined as a sequence of relations $r_1 \rightarrow r_2 \rightarrow ... \rightarrow r_n$, connecting a subject $s$ and an object $o$. The concept of the algorithm is to use these paths as features to decide whether a given triple is noisy or not. Using pruning and other heuristics, specific paths are kept per relation to indicate whether a triple is erroneous. In the end, a confidence score with values [0, 1] is decided for each triple, with low scores indicating noise. 
	
\subsection{TransE}

The basic idea behind the TransE~\cite{transe} model is that, given a triple $(s, r, o)$, the subject $s$ and the relation $r$ can be connected with the object $o$ with low error, meaning $s + r \approx o$. Each component of entity $(s,r,o)$ is represented through an embedding, a vector of fixed size $d$. The fitness of the model is calculated through the energy function: $E(s, r, o) = || s + r - o ||_{l_{2}} $, where $l_{2}$ denotes the $L_2$ norm, and the higher the fitness for a triple, the smaller the value of the energy function. TransE recursively minimizes a pairwise max-margin scoring function that uses the aforementioned energy function and negative sampling for training:
	\begin{dmath}\label{eq:transe_loss}
		Loss = \sum_{\small (s, r, o) \in S}^{}\sum_{ \small (s', r, o') \in S'}{[\gamma + E^+ - E^-]_{+}}
	\end{dmath}
where $E^+$ is the energy function score of a positive triple from the actual KG and $E^- $ is the energy function score of a negative triple  generated by random sampling and $\gamma$ is the hyper-parameter of margin. $[x]_{+}$ denotes the positive part of $x$.

\subsection{Path Ranking Guided Embeddings (PRGE)}

	In the previous subsections, we briefly presented PaTyBRED and TransE. The first, outputs a confidence score for each triple present in the knowledge graph, denoting the probability of the triple being erroneous. The second, generates embeddings for each entity and relation, based on the goodness of fit of the model on the original triples. However, the embedding generation methodology has no inherent way of dealing with knowledge graphs containing errors. As such, an erroneous and a correct triple would both influence the embedding generation procedure the same way, leading to low-quality embeddings directly related to the number of erroneous triples existing in a knowledge graph. This problem exists for all graph embeddings procedures, that cannot intrinsically handle the possibility of encountering noisy input.
	
\begin{figure}
	\centering
    \includegraphics[width=\linewidth]{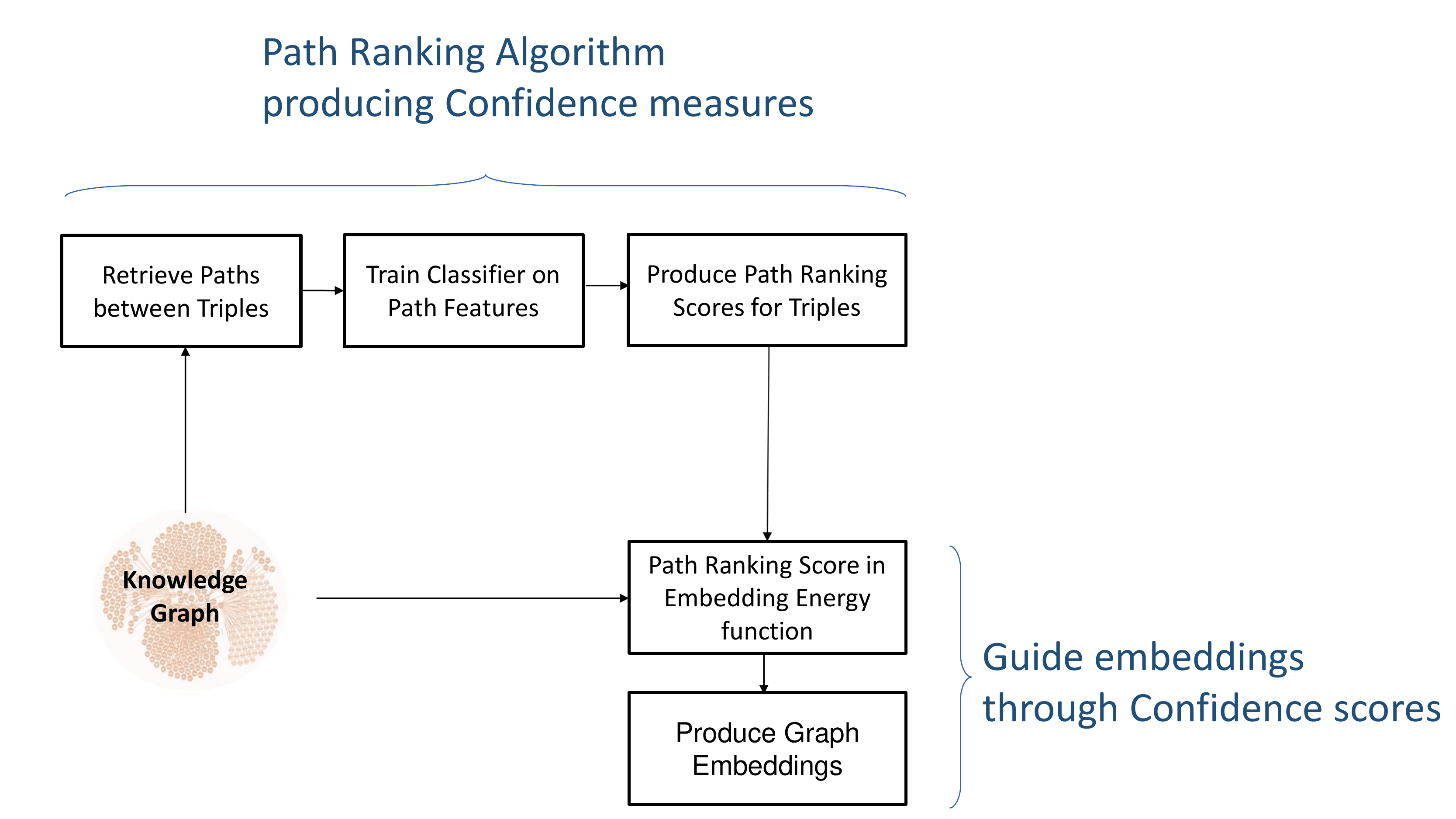}
    \caption{Workflow of the PRGE framework. Applying Path Ranking to calculate confidence scores for each triple and use them to guide the graph embedding procedure.} \label{fig:method_steps}
    \end{figure}

	With the previous concern in mind and taking into account that most real-world knowledge graphs nowadays are noisy due to errors in the automatic construction process, we propose a framework to properly guide the graph embedding creation procedure when working with noisy graphs. In order to showcase this approach, we propose the hybrid approach of Path Ranking Guided Embedding (PRGE). PRGE leverages the PaTyBRED confidence score generated for each triple, to guide the TransE model on focusing on the probably correct triples. An overview of the approach is presented in Figure~\ref{fig:method_steps}.

	In order to do so, we utilize the confidence measure $P(s, r, o)$  of the PaTyBRED method, to lessen the effect that noisy triples have during the training process. This is done by incorporating the confidence score of each triple in Eq. (\ref{eq:transe_loss}) in the following way:
	
	 \begin{dmath}\label{eq:prge_loss}
		Loss = \sum_{(s, r, o) \in S}^{}\sum_{(s', r', o') \in S'}({[\gamma + E^+ - E^-]_{+}}) \cdot P(s, r, o)^{\lambda}
	\end{dmath}
	where $\lambda$ is a scaling parameter to tune the importance of the confidence score. Using this additional exponential weight with the confidence score as a base, we can guide the training procedure to put less emphasis on noisy triples and focus on the probably correct facts in the knowledge graph. This is because noisy triples will have on average a smaller $P(s, r, o)$ than the correct ones, thus decreasing the loss accumulated from noise in Eq. (\ref{eq:prge_loss}) and finally  diminishing the effect the noisy triples have on the generated embedding values.
	
	It is important to note here, that although PRGE is presented here using the PaTyBRED, as the confidence scoring mechanism and TransE,  as the graph embedding procedure, the approach is modular regarding these two components. Any other error detection procedure that generates a score per triple could be plugged in instead of PaTyBRED and likewise, other graph embedding methods could be used instead of TransE. We opted to use the PaTyBRED method to calculate the confidence scores, as it is the most simplified and robust of the PRA-based methods and TransE as it is a well-known, simple yet efficient methodology, that is commonly used as a base for similar approaches. 
	
	As such, PRGE provides a modular framework for combining methodologies from the error detection and graph embedding communities to tackle challenges in real-world applications where noise in knowledge graphs usually exists.

\section{Experiments}\label{sec:exp}

	\subsection{Datasets}\label{subsec:data}
	
	We evaluate the proposed method and other competitors on different tasks related to error detection. We perform experiments on two commonly used KG datasets and a KG created for a real-world application. Statistics for the datasets are presented in Table ~\ref{tab:table_1}. 
	
	The two benchmark datasets are \textit{WN18} from WordNet ~\cite{Wordnet} and \textit{FB15k} from from the FreeBase Knowledge Base ~\cite{Bollacker}. To demonstrate the need for error detection methodologies in real-world applications, we experimented with a Knowledge Graph dataset \textit{(Dementia)} created in the context of the iASiS Project~\cite{krithara2019iasis}. For the needs of the project, we extracted relations between biomedical entities from abstracts of publications related to Dementia in PubMed\footnote{https://www.ncbi.nlm.nih.gov/pubmed/} using automatic tools. Specifically, after fetching abstracts related to Dementia, through semantic MeSH queries\footnote{https://www.nlm.nih.gov/mesh/}, we  use SemRep~\cite{rindflesch2003interaction} for automatically extracting biomedical predications, i.e. semantic triples in the form of subject-predicate-object, from unstructured text. The subject and object  arguments  in  these  predications  are  concepts  from  the  Unified Medical Language System~\cite{bodenreider2004unified} and  the predicate is one of the semantic relations of the Semantic Network~\cite{mccray2003upper}, connecting the semantic types of the subject and object in the context of the specific sentence. 
	
	For constructing the graph, 68,791 publications were fetched using the related MeSH term \textit{Dementia}. More details regarding the exact procedure followed for the creation of this knowledge graph and the extraction process can be found in~\cite{nentidis2019semantic}.

\begin{table}
\centering
 
  \begin{tabular}{p{0.4\linewidth}p{0.1\linewidth}p{0.14\linewidth}p{0.15\linewidth}}
	\toprule
	Dataset & \# Rel & \# Ent & \# Triples \\
	\midrule
	WN18 & 18 & 40,943 & 141,442 \\
	FB15k & 1345 & 14,951 & 483,142 \\
	Dementia  & 64 & 48,008 & 135,000 \\
	\bottomrule
  \end{tabular}
  
  \caption{Details of the 3 datasets used (WordNet, FreeBase and iASiS) in terms of different relation types and size.}
  \label{tab:table_1}
\end{table}

	\subsection{Error Imputation Protocol}\label{subsec:imp}
	
	 To assess the methodologies presented and proposed, we need noise to be present in the KG. However, there are no explicitly-labeled noisy triples in FB15K or WN18. Therefore, we generated new datasets with different percentages of noise levels to simulate real-world knowledge graphs constructed automatically. In order to do so, we construct negative triples following different approaches. The basic idea behind the error imputation process is that for each positive triple (\textit{s, r, o}) in the dataset, we generate a noisy one by corrupting either \textit{s} or \textit{o}. For the FB15K knowledge graph, we follow the procedure described in \cite{CKRL}, where the generation of noise is constrained, in that the new subject $s'$ or object $o'$ should have appeared in the dataset with the same relation $r$. This constraint focuses on generating harder and more confusing noise for any method. On the contrary, negative sampling on WN18 and Dementia KGs was performed randomly, without any constraint, to compare different methods and datasets on different noise types.
	 
	 For each dataset, we generated noisy copies of the original graph with ($10\%$), ($20\%$) and ($40\%$) added noise respectively, to assess the impact of different noise levels in the results. It is also important to note that, all these errors which were imbued to the 3 datasets are labeled as positives for training purposes. This means that the evaluation of the methods will be based on how effective they are in finding these hidden errors in every KG, with no indication of the noise level. 
	 
% \begin{table}[htp]
 
%   \centering
%   \begin{tabular}{p{0.3\linewidth}p{0.15\linewidth}p{0.15\linewidth}p{0.15\linewidth}}
% 	\toprule
% 	Dataset & N1 (10\%) & N2 (20\%) & N3 (40\%) \\
% 	\midrule
% 	WN18 & 14,144 & 28,288 & 56,445 \\
% 	FB15k & 46,408 & 93,782 & 187,925 \\
% 	Dementia & 13,500 & 27,000 & 54,000 \\
% 	\bottomrule
%   \end{tabular}
  
%   \caption{Number of imputed errors based on ratio for each dataset}
%   \label{tab:table_2}
% \end{table} 

	\subsection{Evaluation Protocol} 
	
	Following the same steps as~\cite{Socher}, we compute the energy function $E(s, r, o) = ||s + r - o||$ for each triple in the dataset. Then, we generate a ranking for all triples based on this energy function score.  The smaller the value of the energy value of the triple, the more valid the triple is. As such, we would hope that the erroneous triples would have much greater value than the initial correct ones. To measure this we use the filtered mean rank ($f$MR) and the filtered mean reciprocal rank ($f$MRR)~\cite{PaTyBRED}.
	
% 	\begin{equation}
% 		fMR = \frac{1}{N}\sum_{i = 1}^{N}rank_i - i + 1
% 	\end{equation}
	
% 	\begin{equation}
% 		fMRR = \frac{1}{N}\sum_{i = 1}^{N}\frac{1}{rank_i - i + 1}
% 	\end{equation}

	Additionally, after normalizing the energy function score in the [0, 1] interval, we also use the Area Under the ROC Curve (AUC) to further examine how well algorithms classify the noise as an error. Values close to 0 indicate a correct triple, while values close to 1 indicate an erroneous triple. For $f$MR, lower is better while for $f$MRR and AUC, higher is better.

    \begin{table*}[!htp]
      \centering
     \begin{adjustbox}{width=0.85\linewidth}
      \begin{tabular}{l|ccc|ccc|ccc}
    	\toprule
        \multirow{2}{*}{Dataset} & \multicolumn{3}{c}{WN18-$10\%$} & \multicolumn{3}{c}{WN18-$20\%$} & \multicolumn{3}{c}{WN18-$40\%$} \\
        & {$f$MR} & {$f$MRR} & {AUC} & {$f$MR} & {$f$MRR} & {AUC} & {$f$MR} & {$f$MRR} & {AUC} \\
        \midrule
        PaTyBRED & 4593  & 0.0008 & 0.9673 & 4694  & \textbf{0.0009} & 0.9668 & 4703   & 0.0007 & 0.9668 \\
        \hline
        TransE & 38942 & 0.0002 & 0.7247 & 39339 & 0.0003 & 0.7219 & 44464 & 0.0005 & 0.6857 \\
        PTransE & 45721  & 0.0007 & 0.6768 & 45392  & 0.0003 & 0.6791 & 46412  & 0.0002 & 0.6719 \\
        CKRL & 15738  & \textbf{0.0009} & 0.8887 & 16969  & 0.0007 & 0.8800 & 39253  & \textbf{0.0011} & 0.7225 \\
        \hline
        %PRGE & 9913 & 0.0006 & 0.9299 & 12450  & 0.0004 & 0.9120 & 19956 & 0.0004 & 0.8589 \\
        PRGE & \textbf{3681} &\textbf{ 0.0009} & \textbf{0.9740} & \textbf{3870} & \textbf{0.0009} & \textbf{0.9727} & \textbf{3673} & 0.0008 & \textbf{0.9740} \\
        \bottomrule
      \end{tabular}
      \end{adjustbox}
      
       \caption{Error Detection results for WN18 (Imputing Random Errors).}
       \label{tab:table_3}
    \end{table*}

%%%%  WITHOUT PERCENTS %%%%
\begin{table*}[!htp]
  \centering
   \begin{adjustbox}{width=0.85\linewidth}
  \begin{tabular}{l|ccc|ccc|ccc}
	\toprule
    \multirow{2}{*}{Dataset} & \multicolumn{3}{c}{FB15K-$10\%$} & \multicolumn{3}{c}{FB15K-$20\%$} & \multicolumn{3}{c}{FB15K-$40\%$} \\
    & {$f$MR} & {$f$MRR} & {AUC} & {$f$MR} & {$f$MRR} & {AUC} & {$f$MR} & {$f$MRR} & {AUC} \\
    \midrule
    PaTyBRED & \textbf{41785} & 0.0005 & \textbf{0.9064} & \textbf{46046} & 0.0003 & \textbf{0.8907} & \textbf{53320}  & \textbf{0.0002} & \textbf{0.8694} \\
    \hline
    TransE & 127940 & 0.0002 & 0.7352 & 133763  & 0.0001 & 0.7231 & 169488 & 0.0000 & 0.6492 \\
    PTransE & 166349 & 0.0000 & 0.6557 & 167997  & 0.0000 & 0.6523 & 173643  & 0.0000 & 0.6406 \\
    CKRL & 96113 & 0.0001 & 0.8011 & 101583  & 0.0001 & 0.7897 & 112325  & 0.0001 & 0.7675 \\
    \hline
    
    %PRGE & 89058 & 0.0004 & 0.8157 & 103167 & 0.0002 & 0.7865 & 106907  & 0.0001 & 0.7787 \\
    
    PRGE & 73994  & \textbf{0.0006} & 0.8469 & 89164  & \textbf{0.0005} & 0.8155 & 86347 & \textbf{0.0002} & 0.8213 \\
    \bottomrule
  \end{tabular}
  \end{adjustbox}
  
  \caption{Error Detection results for FB15K (Imputing with Same Relation Errors constrained approach).}
  \label{tab:table_4}
\end{table*}

	\subsection{Comparison and Settings}

    We compare our methodology with the CKRL~\cite{CKRL} and the PTransE~\cite{ptranse} methods. Both of these methodologies are state-of-the-art hybrid methods that also generate embeddings. The CKRL method is build with the error detectiong task in mind and follows a similar procedure for generating the embeddings. However, it uses a complex way of generating the confidence score for each triple, taking into account local and global path confidence scores and a final weighting scheme. The PTransE method is a predecessor to the CKRL method using paths to guide embeddings, albeit in a different way than CKRL. We also provide results using the vanilla PaTyBRED and TransE models on the same tasks, with PaTyBRED being the only non-embedding-generating method.
    
    For all methods, we used the settings and parameters suggested by the corresponding authors on the two benchmark datasets. For consistency, we also used these parameters on the Dementia dataset. In all embeddings methods (including our own implementation) we used $d = 50$ as the size of the dimensions. In CKRL the authors also state that $d = 50$ is the best value for the FB15k dataset. For all datasets, training was limited to 1000 epochs, as further training didn't improve performance substantially. Early stopping was used to determine the best model during these epochs. Regarding the scaling value $\lambda$ of the PRGE method, we use $\lambda = 5$, which yielded the best results on all datasets, after searching over a small subset of possible values. 
    
    %In the following section, PRGE-Scaled refers to a model with $\lambda = 5$ and PRGE refers to a model with $\lambda = 1$ (i.e. unscaled).
	
	\subsection{Results and Discussion}

    \subsubsection*{Error Detection Experiments}
    
	Tables ~\ref{tab:table_3},~\ref{tab:table_4} and~\ref{tab:table_5} demonstrate the results of all approaches for the error detection task on all datasets. Some interesting observations and insights stemming from the results are presented here:

	1) \underline{WN18 Dataset}: Regarding the WN18 dataset, it is evident from Table \ref{tab:table_3} that our proposed PRGE outperforms all other methods. CKRL and PaTyBRED perform similarly on the $f$MRR metric but are outperformed on the other evaluation metrics. It is also evident that, compared to the other methods, PRGE performs better as the noise ratio goes up, with $f$MR and AUC score values being non-decreasing when noise increases from $10\%$ to $40\%$.
	
	\begin{comment}
	
	    Regarding the base methods, PaTyBRED performs better than almost any base embedding method in error detection. Using paths as features to decide whether a triple is noisy or not, gives remarkably good results on all cases. The only exception is our PRGE-Scaled method, which outperforms PaTyBRED on most cases and datasets. 
	
	2) Another interesting result is that PTransE performs worse than TransE on error detection on both datasets. This is also mentioned in \cite{CKRL}, and we confirm it here by using the same energy function ($||s + r - o||$) for all embeddings methods. This is rather unexpected, as PTransE uses additional path information to guide embeddings similar to CKRL, which in turn performs better in all aspects than both TransE and PTransE.
	
	\end{comment}
	
	2) \underline{FB15k Dataset}: Here, PaTyBRED performs better than almost any base embedding method in error detection, indicating that potentially important factors here are the dataset size (see Table \ref{tab:table_1}) and the different error imputation method. However, our PRGE method fares better on the $f$MRR metric, indicating that it separates better the obvious erroneous triples from the others. In addition, our PRGE performs better than all other embedding-based methods. 
	
	\begin{comment}
	
	3) The proposed method (both scaled and unscaled) performs better than CKRL on all datasets, for almost all noise ratios and noise imputation methods. This confirms that the PaTyBRED score (and possibly more PRA variants) as a confidence score is more expressive than the CKRL confidence score in guiding the loss function to produce better embeddings for error detection.

    \end{comment}
    
    3) \underline{Dementia Dataset}: Firstly, as discussed in previous paragraphs, the knowledge graph is very sparse and this dataset had noise present even before the noise imputation, due to the automatic creation process. As such, the actual noise level is much higher than the other datasets. Hence, it poses a far more challenging error detection task. This can be seen in Table \ref{tab:table_5} where all methods perform worse than the other two dataset. In spite of PaTyBRED being slightly better at the ranking metrics, our PRGE method achieves better AUC scores, indicating that on average it can perform better than other models when comparing between an actual and a noisy triple. It can also scale better with the increasing noise ratio, something also seen in the WN18 dataset. In the $40\%$ dataset, our method can achieve better $f$MR score than every method, indicating that in the presence of much noise, something expected in most automatically generated KGs, it can fair better than the competition.
    
    4) \underline{Effect of noise}: As expected, when the noise level rises from $10\%$ to $40\%$, the performance of all models deteriorates regardless of the dataset. However, our model is the most robust, especially when compared to the other embedding methods, showing much smaller fluctuations in performance.

%%%%  WITHOUT PERCENTS %%%%
\begin{table*}[!htp]
\centering
  \begin{adjustbox}{width=0.85\linewidth}
  \begin{tabular}{l|ccc|ccc|ccc}
	\toprule
    \multirow{2}{*}{Dataset} & \multicolumn{3}{c}{Dementia-$10\%$} & \multicolumn{3}{c}{Dementia-$20\%$} & \multicolumn{3}{c}{Dementia-$40\%$} \\
    & {$f$MR} & {$f$MRR} & {AUC} & {$f$MR} & {$f$MRR} & {AUC} & {$f$MR} & {$f$MRR} & {AUC} \\
    \midrule
    PaTyBRED & \textbf{56485} & \textbf{0.0006} & 0.5674 & \textbf{55749} & \textbf{0.0007} & 0.5604 & 59817 & \textbf{0.0003} & 0.5552 \\
    \hline
    TransE & 58014 & 0.0001 & 0.5702 & 59421  & 0.0001 & 0.5599 & 59835  & 0.0000 & 0.5568 \\
    PTransE & 59718  & 0.0002 & 0.5576 & 61518  & 0.0001 & 0.5443 & 65533  & 0.0000 & 0.5146 \\
    CKRL & 60584 & 0.0001 & 0.5512 & 61034  & 0.0001 & 0.5479 & 61089  & 0.0001 & 0.5475 \\
    \hline
    %PRGE & 58049 & 0.0001 & 0.5700 & 58510  & 0.0001 & 0.5666 & 59844  & 0.0002 & 0.5567 \\
    PRGE & 57642  & 0.0001 & \textbf{0.5730} & 58258 & 0.0001 & \textbf{0.5685} & \textbf{59314}  & 0.0001 & \textbf{0.5606} \\
    \bottomrule
  \end{tabular}
  \end{adjustbox}
  
  \caption{Error Detection results for Dementia (Imputing Random Errors).}
  \label{tab:table_5}
\end{table*}

%	5) \underline{PRGE scaling effect}: Regarding our proposed method, we can see that the $\lambda$ scaled PRGE method works better than the unscaled method. This is true for all different noise imputation ratios and datasets, reflecting the importance of weighting the confidence score for each triple during training. 

	5) \underline{PTransE performance}: Another interesting result is that PTransE performs worse than TransE on error detection on all datasets. This is rather unexpected, as PTransE uses additional path information to guide embeddings similar to CKRL, which in turn performs better in all aspects than both TransE and PTransE.
	
	At this point it is important to stress two main advantages of the proposed methodology, alongside the superior performance results:
	\begin{itemize}
	    \item[$\bullet$] \underline{Robust Embeddings}: Contrary to the PRA methods where only a confidence score for each triple is provided in the end, the PRGE method produces embeddings trained and guided by this confidence score. This results in embeddings robust to the inherent noise of the knowledge graph, which can be further used in other embedding-based downstream tasks, such as link prediction, triple classification, clustering, etc.	    \item[$\bullet$] \underline{Modularity}: The proposed PRGE method is agnostic of the underlying energy function and triple-scoring mechanism. Since the path ranking score only acts in a multiplicative manner to the energy function, one can deduce that other embedding energy functions can be used to improve performance. Specifically, methods like TransH and TransR or even Semantic Matching methods like RESCAL can also be used in conjunction with the path ranking score and could improve results. Respectively, instead of the PaTyBRED score, other scoring mechanisms could be employed as well. This makes the proposed PRGE method generic and flexible, allowing for different combinations of techniques for the energy function and the confidence score that could enhance results for the task at hand.
	\end{itemize}

\subsubsection*{Triple Classification Experiments}
To prove the usefulness of robust-to-noise embeddings in downstream tasks, we also performed a triple classification experiment following~\cite{Socher}. Triple classification as a task revolves around predicting whether a triple belongs to a graph or not.  Ultimately, our goal is to predict correct facts in the form of relations $(s,r,o)$ from the test data, using the score $||s + r - o||$ by utilizing the embeddings generated from the various models.

More specifically, to classify whether a triple is valid on not, a threshold $\tau_r$ for every relation $r$ is introduced. This threshold is chosen based on performance on the validation set. The threshold value for every relation is the one that maximizes the classification accuracy on the validation set. Then, using these thresholds, the performance of the model is estimated on the test set.

\begin{table}[htp]
\centering
 
  \begin{tabular}{p{0.3\linewidth}p{0.15\linewidth}p{0.15\linewidth}p{0.15\linewidth}}
	\toprule
	Dataset  & $10\%$ & $20\%$ & $40\%$ \\
	\midrule
	TransE & \textbf{0.717} & 0.703 & 0.671 \\
	PTransE & 0.686 & 0.678 & 0.670 \\
	CKRL & 0.639 & 0.709 & 0.691 \\
	\hline
	%PRGE & 0.712 & \textbf{0.715} & 0.681 \\
	%PRGE-Scaled & 0.715 & 0.712 & \textbf{0.702} \\
	PRGE & 0.715 & \textbf{0.715} & \textbf{0.702} \\
	\bottomrule
  \end{tabular}
  
  \caption{Triple Classification on the FB15k with the three different noise ratios. The scores correspond to ROC-AUC.}
  \label{tab:table_triple_classification}
\end{table}

Results on the FB15k, for different noise ratios as before, can be seen in Table~\ref{tab:table_triple_classification}. We can see that as the noise ratio gets larger, PRGE performs better than the other methods. In addition, the PRGE method consistently outperforms CKRL and PTransE on all noise ratios, indicating that using the path ranking score to train embeddings yields better results. The same behavior was observed on the Dementia and WN18 datasets on average, but they were omitted for brevity. Conclusively, we can see that utilizing the PRGE framework to incorporate an error estimation score during the training process of the embeddings, actually helps in other downstream tasks with the generation of noise-robust embeddings.

\begin{table*}[!htp]
    
  %\centering
  \begin{tabular}{p{0.3\linewidth}p{0.15\linewidth}p{0.44\linewidth}}
	\toprule
    Erroneous triple (s - r - o) & Type of error & Supporting text\\
    \midrule
    acetonitrile - AUGMENTS - 80\% & Extraction Error & \textit{... The \textbf{acetonitrile} concentration was increased to 80\% in 5 min and then held in \textbf{80\%} acetonitrile for an additional 5 min ...}\\
    % Haiti - PART_OF - Cells	& \textit{... Cos7 Cell Line Experiments HA-NMNAT2 was expressed in Cos7 cells using lipofectamine ...}\\
    Denmark - PART\_OF - Neurons	& Extraction Error & \textit{... \textbf{Neurons} transfected with \textbf{DA}-GFP were found to have dendritic spines that had significantly lengthened necks compared to ...}\\
    %\hline
    % Patients - LOCATION\_OF - glutamic acid	& TG & \textit{... A significantly increased frequency of the \textbf{Glu/Glu genotype} in late onset AD LOAD \textbf{patients} was found ...}\\
    %\hline
    Cells - PART\_OF - Medial geniculate body & Too General &\textit{... Dendritic spines of the polyhedral and elongated \textbf{cells} of the \textbf{medial geniculate bodies} were decreased in number ...}\\
    %\hline
    % Entire thumb - PART\_OF - Patients & Too General & \textit{... The \textbf{patient} is asked to hold the ruler with his \textbf{thumb} and forefinger and to release the ruler while the investigator continues to ...}\\
    DDMS - PART\_OF - Homo sapiens & Too Specific & \textit{... HT-22 hippocampal cells and confirms observations using brain extracts from monkey, mouse, rat and \textbf{human} \textbf{DDM} ...}\\
    %\hline
    Brain - LOCATION\_OF - Decreased plasmalogens & Too Specific & \textit{... data suggest that long-term alterations in plasmalogen synthesis degradation result in \textbf{decreased brain plasmalogen} levels, a hallmark feature of AD ...}\\
    %\hline
    \bottomrule
  \end{tabular}
  
  \caption{The lowest-scoring triples of the erroneous categories, two from each one, along with the type of error  and the initial text they were extracted from.}
  \label{tab:table_errors}
\end{table*}

\subsubsection*{Qualitative Results on the Dementia Dataset}
Since our final goal is to detect the erroneous triples already present in a Knowledge Graph, we also performed a qualitative analysis of the predictions given by the model. We are interested in seeing if the lowest scoring triples, the ones that our model deemed most probable to be errors, should be removed. We focused on the Dementia dataset to showcase what the PRGE model predicts as erroneous in a real-world application. The PRGE model was trained on the initial Dementia dataset, before the imputation process. Thus, we are trying to detect the actual noise present in the Knowledge Graph. To assess the validity of the predictions we devised the following annotation task:

\begin{figure}[!hp]
\includegraphics[width=\linewidth]{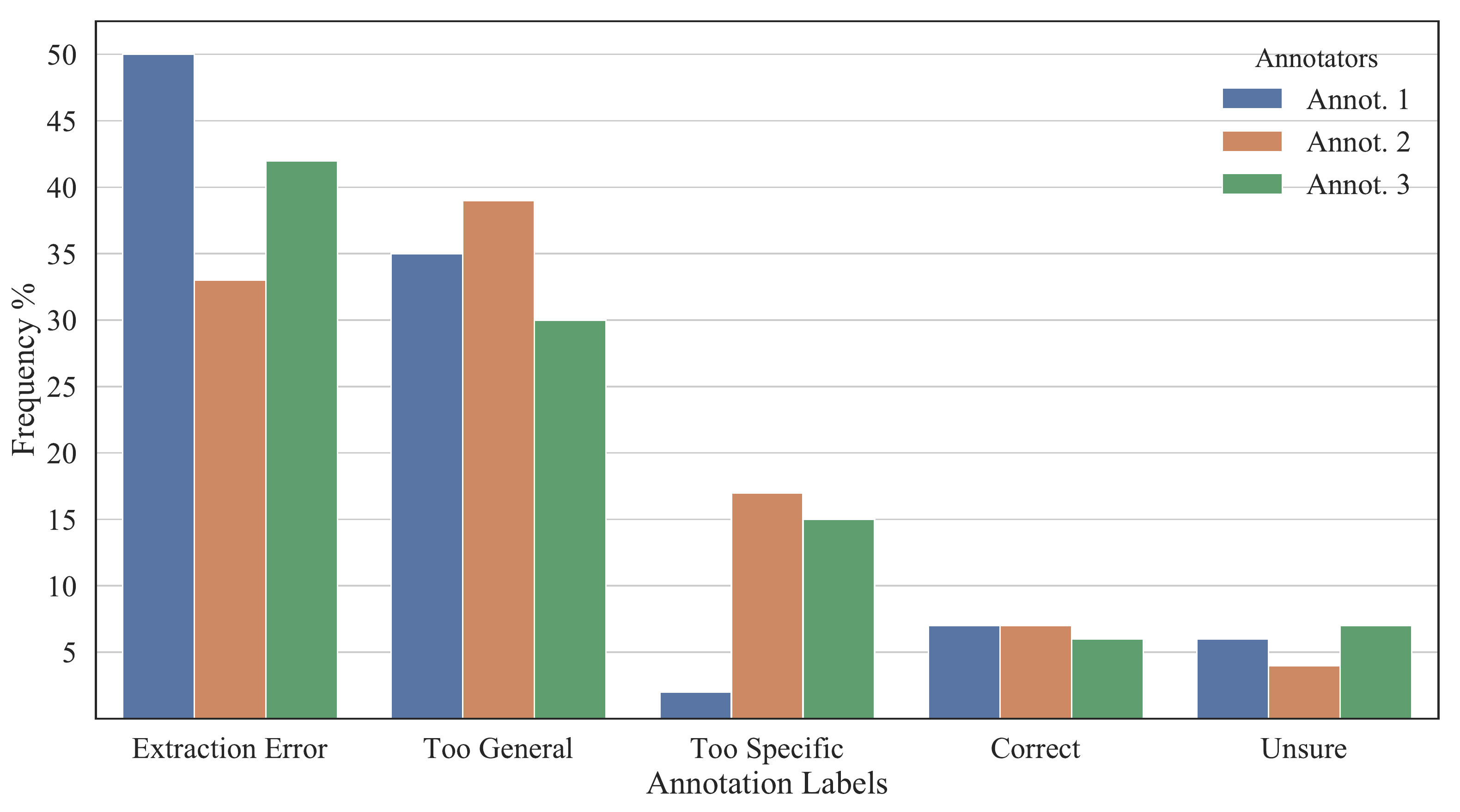}
\caption{Manual evaluation of the top 100 errors, according to the PRGE model, from the Dementia Dataset.} \label{fig:annotation_results}
\end{figure}

Firstly, we fetched the top-100 lowest-scoring triples, as predicted by our model. We also fetched the exact textual snippets from the publications that these triples were found in. Three human experts on natural language processing and bioinformatics were presented with these triples alongside their corresponding text. The triples were presented in the form (Entity1 – Relation – Entity2). They were asked to assess the quality of the triple given the corresponding textual content and how useful was the piece of information that was extracted. Specifically, the annotators could select one of the following labels for each triple: ``\textit{Correct}'', ``\textit{Unsure}'', ``\textit{Extraction Error}'', ``\textit{Too General}'' and ``\textit{Too Specific}''.

The last three labels “\textit{Extraction Error}”, “\textit{Too Specific}” and “\textit{Too General}” are all errors according to this evaluation scheme. We devised multiple labels for the errors because it is important to have a qualitative analysis of the errors made. For example, high “\textit{Extraction Errors}” would indicate errors made from the relation extraction tool working directly on the text and would support research towards enhancing that part of the pipeline to reduce the error propagation. On the other hand, higher “\textit{Too General}/\textit{Too Specific}” errors would insinuate that the relation extraction tool works correctly, however, the extracted triples are not important for the task at hand. In that case, we could devise and apply a post-processing step to keep meaningful triples.

The results of the annotator’s evaluation can be seen in Figure~\ref{fig:annotation_results}. We can see that for all annotators more than $85\%$ of these triples seemed to be erroneous given the context. On the other hand, out of these lowest-scoring triples only $\approx8\%$ is actually correct, across all annotators. This indicates the high precision of the prediction and allows us to be fairly confident on the scoring of the model.

As an added example, some of the manually assessed triples can be seen in Table~\ref{tab:table_errors}. There, we present some of lowest-scoring triples from each one of the erroneous categories and the corresponding text they were derived from. The triples presented were all unanimously labeled as the specific type of error from the annotators. As we can see, the triple \textit{Denmark - PART\_OF - Neurons} is an exemplary case where the extraction tool failed to identify the correct entities and their relation. On the other hand, the triple \textit{Cells - PART\_OF - Medial geniculate body} may be correctly extracted from the text but provides little to no added value to the aggregated knowledge. Having multiple such facts deteriorates the final quality, as each model has to capture a lot of ``useless'' information. 
%Thus, these examples showcase the importance of the discrimination regarding the type of error made, as well as, the added value of performing such an analysis in noisy graphs.

\subsection{Limitations}
This research, however, is subject to some limitations. The first lies in the marginal improvements presented. The PRGE method indeed performs better across datasets over all embedding-based methods but in most cases, the difference is rather small. Also, PaTyBRED is used as a strong baseline for error detection and as seen it outperforms most embedding-based methods in many cases, including ours in some. The downside with PaTyBRED though, is that it does not generate embeddings as a by-product for further downstream tasks, while PRGE does. Nonetheless, more experimentation and a thorough comparison with other state-of-the-art methods in error detection is needed, also taking into account probabilistic approaches and rule-based techniques as presented in the Related Work section.

Secondly, the technical novelty presented is limited to the combination of two well-known methodologies. TransE and PaTyBRED were used as the core components of the PRGE method to showcase this modular paradigm of combining such methodologies. However, to establish the PRGE framework as a useful way of combining graph-embedding techniques with error detection methods to handle inherent noise in KGs, an extended study with different combination of methodologies is needed. Moreover, the PRGE-generated embeddings of each such combination should be tested against the original ones in different downstream tasks to actually prove their advantages and robustness to noise.

\section{Conclusion}\label{sec:concl_fut}

	In this paper, we compared different methods for error detection in knowledge graphs and proposed a modular framework for combining them. The proposed PRGE method combines the path ranking score of a triple with the graph embedding framework, creating embeddings robust towards to noise present in the graph. We assessed both the quantitative and qualitative performance of the framework. Utilizing the score from the best-performing PRA-based method (PaTyBRED) to train the embeddings, we have managed to overcome other state-of-the-art hybrid methods (PTransE, CKRL) on all datasets and enhance the classification results of PaTyBRED on two of the three datasets. Moreover, by combining these two frameworks we extended the PRA methods with the ability to generate embeddings for the entities, that could be used further. We also performed a qualitative evaluation of the possible errors detected in a real-world dataset, to showcase the importance of such approaches in actual applications. We then discussed the limitations of the present study and possible future work to mitigate them. Overall, we have proposed a generic framework to generate embeddings resilient to noise and we proved that they can also be used in multiple downstream tasks enhancing performance in the presence of noise\footnote{Code and material can be found here: https://github.com/RomFas/PRGE/}. 
	
% 	As future work, a plethora of other combinations of core-components could be also explored. In the future, the following directions will be researched: Firstly, energy functions from other embedding methods, such as TransR~\cite{transr} will be used. Secondly, we will try to further expand the PRGE-Scaled method by including an additional parameter $\theta$ to the pairwise max-margin part of the loss function. Lastly, inspired by rule-guided embedding frameworks such as KALE~\cite{kale} and RUGE~\cite{RUGE}, we plan to combine path ranking scores with rule-based scores or rule-based training to enhance the results on the error detection task and propose a more unifying framework.
	
\section*{Acknowledgement}

This paper is supported by European Union's Horizon 2020 research and innovation programme under grant agreement No. 727658, project IASIS (Integration and analysis of heterogeneous big data for precision medicine and suggested treatments for different types of patients).

\bibliography{bibliography.bib}

\begin{thebibliography}{39}
\providecommand{\natexlab}[1]{#1}
\providecommand{\url}[1]{\texttt{#1}}
\providecommand{\urlprefix}{URL }
\expandafter\ifx\csname urlstyle\endcsname\relax
  \providecommand{\doi}[1]{doi:\discretionary{}{}{}#1}\else
  \providecommand{\doi}{doi:\discretionary{}{}{}\begingroup
  \urlstyle{rm}\Url}\fi

\bibitem[{Abedini, Keyvanpour, and Menhaj(2020)}]{abedini2020correction}
Abedini, F.; Keyvanpour, M.~R.; and Menhaj, M.~B. 2020.
\newblock Correction Tower: A general embedding method of the error recognition
  for the knowledge graph correction.
\newblock \emph{International Journal of Pattern Recognition and Artificial
  Intelligence} 2059034.

\bibitem[{Akoglu, Tong, and Koutra(2015)}]{akoglu2015graph}
Akoglu, L.; Tong, H.; and Koutra, D. 2015.
\newblock Graph based anomaly detection and description: a survey.
\newblock \emph{Data mining and knowledge discovery} 29(3): 626--688.

\bibitem[{Auer et~al.(2007)Auer, Bizer, Kobilarov, Lehmann, Cyganiak, and
  Ives}]{DBPedia}
Auer, S.; Bizer, C.; Kobilarov, G.; Lehmann, J.; Cyganiak, R.; and Ives, Z.
  2007.
\newblock DBpedia: A Nucleus for a Web of Open Data.
\newblock In Aberer, K.; Choi, K.-S.; Noy, N.; Allemang, D.; Lee, K.-I.; Nixon,
  L.; Golbeck, J.; Mika, P.; Maynard, D.; Mizoguchi, R.; Schreiber, G.; and
  Cudr{\'e}-Mauroux, P., eds., \emph{The Semantic Web}, 722--735. Berlin,
  Heidelberg: Springer Berlin Heidelberg.
\newblock ISBN 978-3-540-76298-0.

\bibitem[{Belth et~al.(2020)Belth, Zheng, Vreeken, and
  Koutra}]{belth2020normal}
Belth, C.; Zheng, X.; Vreeken, J.; and Koutra, D. 2020.
\newblock What is Normal, What is Strange, and What is Missing in a Knowledge
  Graph: Unified Characterization via Inductive Summarization.
\newblock In \emph{Proceedings of The Web Conference 2020}, 1115--1126.

\bibitem[{Bodenreider(2004)}]{bodenreider2004unified}
Bodenreider, O. 2004.
\newblock The unified medical language system (UMLS): integrating biomedical
  terminology.
\newblock \emph{Nucleic acids research} 32(suppl\_1): D267--D270.

\bibitem[{Bollacker et~al.(2008)Bollacker, Evans, Paritosh, Sturge, and
  Taylor}]{Bollacker}
Bollacker, K.; Evans, C.; Paritosh, P.; Sturge, T.; and Taylor, J. 2008.
\newblock Freebase: A Collaboratively Created Graph Database for Structuring
  Human Knowledge.
\newblock In \emph{Proceedings of the 2008 ACM SIGMOD International Conference
  on Management of Data}, SIGMOD '08, 1247--1250. New York, NY, USA: ACM.
\newblock ISBN 978-1-60558-102-6.
\newblock \doi{10.1145/1376616.1376746}.
\newblock \urlprefix\url{http://doi.acm.org/10.1145/1376616.1376746}.

\bibitem[{Bordes et~al.(2013)Bordes, Usunier, Garcia-Duran, Weston, and
  Yakhnenko}]{transe}
Bordes, A.; Usunier, N.; Garcia-Duran, A.; Weston, J.; and Yakhnenko, O. 2013.
\newblock Translating Embeddings for Modeling Multi-relational Data.
\newblock In Burges, C. J.~C.; Bottou, L.; Welling, M.; Ghahramani, Z.; and
  Weinberger, K.~Q., eds., \emph{Advances in Neural Information Processing
  Systems 26}, 2787--2795. Curran Associates, Inc.
\newblock
  \urlprefix\url{http://papers.nips.cc/paper/5071-translating-embeddings-for-modeling-multi-relational-data.pdf}.

\bibitem[{Cai, Zheng, and Chang(2017)}]{survey2}
Cai, H.; Zheng, V.~W.; and Chang, K.~C. 2017.
\newblock A Comprehensive Survey of Graph Embedding: Problems, Techniques and
  Applications.
\newblock \emph{CoRR} abs/1709.07604.
\newblock \urlprefix\url{http://arxiv.org/abs/1709.07604}.

\bibitem[{Carlson et~al.(2010)Carlson, Betteridge, Wang, Hruschka, and
  Mitchell}]{NELL}
Carlson, A.; Betteridge, J.; Wang, R.~C.; Hruschka, Jr., E.~R.; and Mitchell,
  T.~M. 2010.
\newblock Coupled Semi-supervised Learning for Information Extraction.
\newblock In \emph{Proceedings of the Third ACM International Conference on Web
  Search and Data Mining}, WSDM '10, 101--110. New York, NY, USA: ACM.
\newblock ISBN 978-1-60558-889-6.
\newblock \doi{10.1145/1718487.1718501}.
\newblock \urlprefix\url{http://doi.acm.org/10.1145/1718487.1718501}.

\bibitem[{Cheng et~al.(2018)Cheng, Chen, Yuan, and Wang}]{cheng2018rule}
Cheng, Y.; Chen, L.; Yuan, Y.; and Wang, G. 2018.
\newblock Rule-based graph repairing: Semantic and efficient repairing methods.
\newblock In \emph{2018 IEEE 34th International Conference on Data Engineering
  (ICDE)}, 773--784. IEEE.

\bibitem[{Fellbaum(2005)}]{Wordnet}
Fellbaum, C. 2005.
\newblock WordNet and Wordnets.
\newblock In Barber, A., ed., \emph{Encyclopedia of Language and Linguistics},
  2--665. Elsevier.

\bibitem[{Gardner and Mitchell(2015)}]{SFE}
Gardner, M.; and Mitchell, T. 2015.
\newblock Efficient and Expressive Knowledge Base Completion Using Subgraph
  Feature Extraction.
\newblock In \emph{Proceedings of the 2015 Conference on Empirical Methods in
  Natural Language Processing}, 1488--1498. Lisbon, Portugal: Association for
  Computational Linguistics.
\newblock \doi{10.18653/v1/D15-1173}.
\newblock \urlprefix\url{https://www.aclweb.org/anthology/D15-1173}.

\bibitem[{Guo et~al.(2017)Guo, Wang, Wang, Wang, and Guo}]{guo2017knowledge}
Guo, S.; Wang, Q.; Wang, L.; Wang, B.; and Guo, L. 2017.
\newblock Knowledge graph embedding with iterative guidance from soft rules.
\newblock \emph{arXiv preprint arXiv:1711.11231} .

\bibitem[{Heindorf et~al.(2016)Heindorf, Potthast, Stein, and
  Engels}]{Wiki_vandalism}
Heindorf, S.; Potthast, M.; Stein, B.; and Engels, G. 2016.
\newblock Vandalism detection in wikidata.
\newblock In \emph{Proceedings of the 25th ACM International on Conference on
  Information and Knowledge Management}, 327--336. ACM.

\bibitem[{Ho et~al.(2018)Ho, Stepanova, Gad-Elrab, Kharlamov, and
  Weikum}]{ho2018rule}
Ho, V.~T.; Stepanova, D.; Gad-Elrab, M.~H.; Kharlamov, E.; and Weikum, G. 2018.
\newblock Rule learning from knowledge graphs guided by embedding models.
\newblock In \emph{International Semantic Web Conference}, 72--90. Springer.

\bibitem[{Hong, Bu, and Jiang(2020)}]{hong2020rule}
Hong, Y.; Bu, C.; and Jiang, T. 2020.
\newblock Rule-enhanced Noisy Knowledge Graph Embedding via Low-quality Error
  Detection.
\newblock In \emph{2020 IEEE International Conference on Knowledge Graph
  (ICKG)}, 544--551. IEEE.

\bibitem[{Jia et~al.(2018)Jia, Xiang, Chen, and E}]{kgttm}
Jia, S.; Xiang, Y.; Chen, X.; and E, S. 2018.
\newblock {TTMF:} {A} Triple Trustworthiness Measurement Frame for Knowledge
  Graphs.
\newblock \emph{CoRR} abs/1809.09414.
\newblock \urlprefix\url{http://arxiv.org/abs/1809.09414}.

\bibitem[{Jia et~al.(2019)Jia, Xiang, Chen, and Wang}]{jia2019triple}
Jia, S.; Xiang, Y.; Chen, X.; and Wang, K. 2019.
\newblock Triple trustworthiness measurement for knowledge graph.
\newblock In \emph{The World Wide Web Conference}, 2865--2871.

\bibitem[{Kadlec, Bajgar, and Kleindienst(2017)}]{kadlec2017knowledge}
Kadlec, R.; Bajgar, O.; and Kleindienst, J. 2017.
\newblock Knowledge base completion: Baselines strike back.
\newblock \emph{arXiv preprint arXiv:1705.10744} .

\bibitem[{Krithara et~al.(2019)Krithara, Aisopos, Rentoumi, Nentidis,
  Bougatiotis, Vidal, Menasalvas, Rodriguez-Gonzalez, Samaras, Garrard
  et~al.}]{krithara2019iasis}
Krithara, A.; Aisopos, F.; Rentoumi, V.; Nentidis, A.; Bougatiotis, K.; Vidal,
  M.-E.; Menasalvas, E.; Rodriguez-Gonzalez, A.; Samaras, E.; Garrard, P.;
  et~al. 2019.
\newblock iASiS: Towards Heterogeneous Big Data Analysis for Personalized
  Medicine.
\newblock In \emph{2019 IEEE 32nd International Symposium on Computer-Based
  Medical Systems (CBMS)}, 106--111. IEEE.

\bibitem[{Lao and Cohen(2010)}]{PRA}
Lao, N.; and Cohen, W.~W. 2010.
\newblock Relational retrieval using a combination of path-constrained random
  walks.
\newblock \emph{Machine Learning} 81(1): 53--67.
\newblock ISSN 1573-0565.
\newblock \doi{10.1007/s10994-010-5205-8}.
\newblock \urlprefix\url{https://doi.org/10.1007/s10994-010-5205-8}.

\bibitem[{Lehmann et~al.(2012)Lehmann, Gerber, Morsey, and Ngonga}]{Defacto}
Lehmann, J.; Gerber, D.; Morsey, M.; and Ngonga, Ngomo, A.-C. 2012.
\newblock Defacto - Deep Fact Validation.
\newblock In \emph{The Semantic Web -- ISWC 2012}, 312 -- 327. Berlin,
  Heidelberg: Springer Berlin Heidelberg.

\bibitem[{Lin et~al.(2015)Lin, Liu, Luan, Sun, Rao, and Liu}]{ptranse}
Lin, Y.; Liu, Z.; Luan, H.; Sun, M.; Rao, S.; and Liu, S. 2015.
\newblock Modeling Relation Paths for Representation Learning of Knowledge
  Bases.
\newblock In \emph{Proceedings of the 2015 Conference on Empirical Methods in
  Natural Language Processing}, 705--714. Lisbon, Portugal: Association for
  Computational Linguistics.
\newblock \doi{10.18653/v1/D15-1082}.
\newblock \urlprefix\url{https://www.aclweb.org/anthology/D15-1082}.

\bibitem[{McCray(2003)}]{mccray2003upper}
McCray, A.~T. 2003.
\newblock An upper-level ontology for the biomedical domain.
\newblock \emph{Comparative and Functional Genomics} 4(1): 80--84.

\bibitem[{Melo and Paulheim(2017)}]{PaTyBRED}
Melo, A.; and Paulheim, H. 2017.
\newblock Detection of Relation Assertion Errors in Knowledge Graphs.
\newblock In \emph{Proceedings of the Knowledge Capture Conference}, K-CAP
  2017, 22:1--22:8. New York, NY, USA: ACM.
\newblock ISBN 978-1-4503-5553-7.
\newblock \doi{10.1145/3148011.3148033}.
\newblock \urlprefix\url{http://doi.acm.org/10.1145/3148011.3148033}.

\bibitem[{Nentidis et~al.(2019)Nentidis, Bougiatiotis, Krithara, and
  Paliouras}]{nentidis2019semantic}
Nentidis, A.; Bougiatiotis, K.; Krithara, A.; and Paliouras, G. 2019.
\newblock Semantic integration of disease-specific knowledge .

\bibitem[{Paulheim(2016)}]{KGR}
Paulheim, H. 2016.
\newblock Knowledge graph refinement: A survey of approaches and methods.
\newblock \emph{Semantic Web} 8: 489--508.
\newblock \doi{10.3233/SW-160218}.

\bibitem[{Paulheim and Bizer(2014)}]{SDValidate}
Paulheim, H.; and Bizer, C. 2014.
\newblock Improving the Quality of Linked Data Using Statistical Distributions.
\newblock \emph{Internation Journal on Semantic Web and Information Systems
  (IJSWIS)} 10(2): 63 -- 86.

\bibitem[{Rashid et~al.(2019)Rashid, Torchiano, Rizzo, Mihindukulasooriya, and
  Corcho}]{rashid2019quality}
Rashid, M.; Torchiano, M.; Rizzo, G.; Mihindukulasooriya, N.; and Corcho, O.
  2019.
\newblock A quality assessment approach for evolving knowledge bases.
\newblock \emph{Semantic Web} 10(2): 349--383.

\bibitem[{Rindflesch and Fiszman(2003)}]{rindflesch2003interaction}
Rindflesch, T.~C.; and Fiszman, M. 2003.
\newblock The interaction of domain knowledge and linguistic structure in
  natural language processing: interpreting hypernymic propositions in
  biomedical text.
\newblock \emph{Journal of biomedical informatics} 36(6): 462--477.

\bibitem[{Shi and Weninger(2016)}]{shi2016discriminative}
Shi, B.; and Weninger, T. 2016.
\newblock Discriminative predicate path mining for fact checking in knowledge
  graphs.
\newblock \emph{Knowledge-based systems} 104: 123--133.

\bibitem[{Socher et~al.(2013)Socher, Chen, Manning, and Ng}]{Socher}
Socher, R.; Chen, D.; Manning, C.~D.; and Ng, A. 2013.
\newblock Reasoning With Neural Tensor Networks for Knowledge Base Completion.
\newblock In Burges, C. J.~C.; Bottou, L.; Welling, M.; Ghahramani, Z.; and
  Weinberger, K.~Q., eds., \emph{Advances in Neural Information Processing
  Systems 26}, 926--934. Curran Associates, Inc.
\newblock
  \urlprefix\url{http://papers.nips.cc/paper/5028-reasoning-with-neural-tensor-networks-for-knowledge-base-completion.pdf}.

\bibitem[{Suchanek, Kasneci, and Weikum(2007)}]{YAGO}
Suchanek, F.~M.; Kasneci, G.; and Weikum, G. 2007.
\newblock Yago: A Core of Semantic Knowledge.
\newblock In \emph{Proceedings of the 16th International Conference on World
  Wide Web}, WWW '07, 697--706. New York, NY, USA: ACM.
\newblock ISBN 978-1-59593-654-7.
\newblock \doi{10.1145/1242572.1242667}.
\newblock \urlprefix\url{http://doi.acm.org/10.1145/1242572.1242667}.

\bibitem[{Tanon et~al.(2016)Tanon, Vrandečić, Schaffert, Steiner, and
  Pintscher}]{Wikidata}
Tanon, T.~P.; Vrandečić, D.; Schaffert, S.; Steiner, T.; and Pintscher, L.
  2016.
\newblock From Freebase to Wikidata: The Great Migration.
\newblock In \emph{World Wide Web Conference}.

\bibitem[{Toutanova and Chen(2015)}]{toutanova2015observed}
Toutanova, K.; and Chen, D. 2015.
\newblock Observed versus latent features for knowledge base and text
  inference.
\newblock In \emph{Proceedings of the 3rd Workshop on Continuous Vector Space
  Models and their Compositionality}, 57--66.

\bibitem[{Wang et~al.(2017)Wang, Mao, Wang, and Guo}]{survey}
Wang, Q.; Mao, Z.; Wang, B.; and Guo, L. 2017.
\newblock Knowledge Graph Embedding: A Survey of Approaches and Applications.
\newblock \emph{IEEE Transactions on Knowledge and Data Engineering} PP: 1--1.
\newblock \doi{10.1109/TKDE.2017.2754499}.

\bibitem[{Xie, Liu, and Sun(2017)}]{CKRL}
Xie, R.; Liu, Z.; and Sun, M. 2017.
\newblock Does William Shakespeare {REALLY} Write Hamlet? Knowledge
  Representation Learning with Confidence.
\newblock \emph{CoRR} abs/1705.03202.
\newblock \urlprefix\url{http://arxiv.org/abs/1705.03202}.

\bibitem[{Zhang et~al.(2019)Zhang, Paudel, Wang, Chen, Zhu, Zhang, Bernstein,
  and Chen}]{zhang2019iteratively}
Zhang, W.; Paudel, B.; Wang, L.; Chen, J.; Zhu, H.; Zhang, W.; Bernstein, A.;
  and Chen, H. 2019.
\newblock Iteratively learning embeddings and rules for knowledge graph
  reasoning.
\newblock In \emph{The World Wide Web Conference}, 2366--2377.

\bibitem[{Zhao, Feng, and Gallinari(2019)}]{zhao2019embedding}
Zhao, Y.; Feng, H.; and Gallinari, P. 2019.
\newblock Embedding Learning with Triple Trustiness on Noisy Knowledge Graph.
\newblock \emph{Entropy} 21(11): 1083.

\end{thebibliography}
\end{document}